\documentclass{article}
\usepackage[T1]{fontenc}
\usepackage[utf8]{inputenc}
\usepackage{iclr2027_conference,times}
\iclrfinalcopy
\usepackage{amsmath,amssymb,graphicx,booktabs,array,tabularx,longtable,microtype,xcolor}
\usepackage{url}
\usepackage[hidelinks]{hyperref}
\usepackage{flafter}
\newcolumntype{Y}{>{\raggedright\arraybackslash}X}
\newcommand{\vectD}{\mathbf D}
\newcommand{\JS}{\operatorname{JS}}
\newcommand{\RMS}{\operatorname{RMS}}
\newcommand{\ind}{\mathbf 1}
\newcommand{\threeplots}[3]{\includegraphics[width=.326\linewidth]{#1}\hfill\includegraphics[width=.326\linewidth]{#2}\hfill\includegraphics[width=.326\linewidth]{#3}}
\newcommand{\twoplots}[2]{\includegraphics[width=.490\linewidth]{#1}\hfill\includegraphics[width=.490\linewidth]{#2}}
\graphicspath{{figures/}}
\hypersetup{pdftitle={Persistent Depth Ordering amid Shifting Block-Bypass Responses in Language Model Pretraining},pdfauthor={Shengye Tao, Yinzhu Cheng, Haihua Xie}}
\title{Persistent Depth Ordering amid\\Shifting Block-Bypass Responses in\\Language Model Pretraining}
\author{Shengye Tao\textsuperscript{1,2}\quad Yinzhu Cheng\textsuperscript{3,2}\quad Haihua Xie\textsuperscript{2}\\[0.5ex]
\normalfont\small\textsuperscript{1}Beijing University of Civil Engineering and Architecture\\
\normalfont\small\textsuperscript{2}Beijing Institute of Mathematical Sciences and Applications (BIMSA)\\
\normalfont\small\textsuperscript{3}Institute of Statistics and Big Data, Renmin University of China}
\begin{document}
\maketitle
\begin{abstract}
Layer interventions are widely used to probe the internal organization of language models, yet most analyses examine a single training checkpoint even though model representations and computations evolve throughout pretraining. This leaves open which depth-dependent intervention responses reflect persistent organization and which are transient consequences of training. We study this question using single-block identity bypass on fixed teacher-forced contexts across five released trajectories and 11 model--domain combinations. We find that block-bypass responses retain recognizable depth ordering while their magnitudes redistribute: nearby checkpoints preserve stronger rank correspondence than distant ones, and large changes concentrate at positions that recur across text samples and transfer across evaluation domains. Controlled experiments further show that changes in the natural bypass effect cannot be reduced to a single downstream sensitivity: in replicated Pythia runs, local missing-update magnitude grows while the pooled matched downstream response decreases, whereas OLMo-2 7B exhibits a different balance. These matched responses also depend on perturbation strength and direction, without identifying targeted compensation. Together, our results show that longitudinal layer sensitivity is structured but not static, and that single-checkpoint intervention responses should be interpreted in the context of how the underlying perturbation pathway evolves during training.
\end{abstract}
\section{Introduction}
Language-model interpretability often examines a single training checkpoint, although the model being interpreted develops throughout pretraining. A similar ordering of sensitive blocks across checkpoints need not mean that their individual responses stay fixed. We ask: \emph{which aspects of depth-dependent organization persist during pretraining, and which change as learning proceeds?} Our central finding is that \emph{depth ordering can remain recognizable even while individual block responses change substantially}.

\begin{figure}[!ht]
\centering\includegraphics[width=\linewidth]{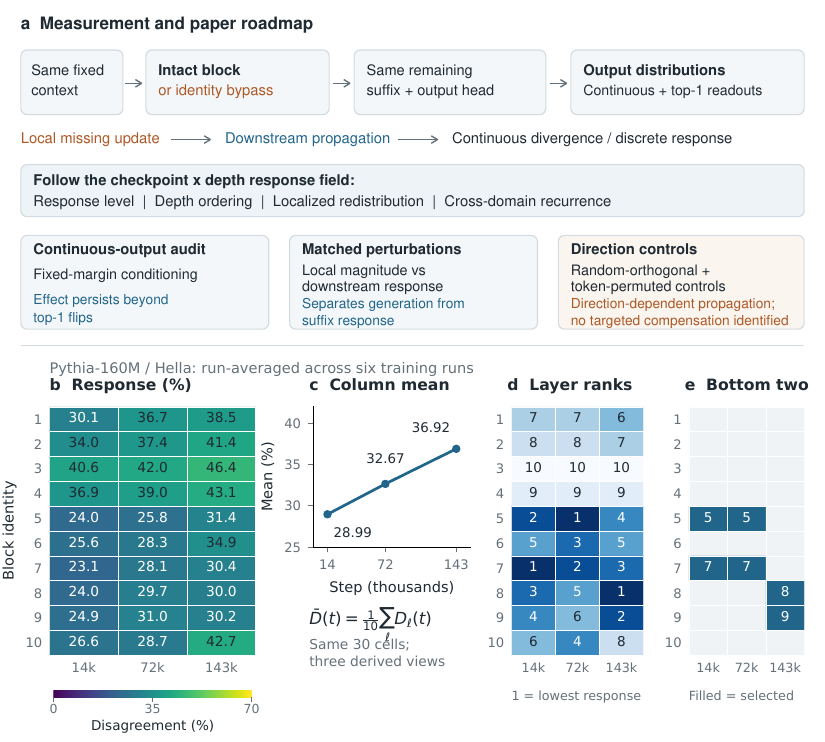}
\caption{\textbf{Measurement and paper roadmap.} (a) Fixed teacher-forced contexts pass through an intact block or identity bypass, then the same suffix and output head. We follow response fields over training, then examine continuous outputs, matched perturbations and direction controls. (b) Historical Pythia-160M Hella responses average six runs over ten interior blocks at 14k, 72k and 143k; entries are percentages. (c--e) Means, ascending ranks and bottom-two selections derive from the same 30 mean-profile cells. Block indices are zero-based. Columns are observed checkpoints, not interpolated intervals.}
\label{fig:workflow}
\end{figure}

Longitudinal studies of attention heads and circuits show how mechanisms emerge and computations persist as their implementations change \citep{olsson2022,nanda2023progress,tigges2024circuits}. Depth offers a direct coordinate for studying persistence without first aligning learned features or circuits: within a fixed architecture, block positions are aligned across checkpoints. We bypass one complete Transformer block at a time and measure the resulting change in next-token predictions on fixed teacher-forced contexts. Repeating this intervention across depth and checkpoints produces a response map. General, mathematical, and code text provide different views of the same trajectory. Figure~\ref{fig:workflow} gives the measurement and paper roadmap.

Nearby checkpoints retain similar sensitivity rankings, while large changes concentrate at a minority of positions and recur across evaluation domains. Persistence thus depends on what is measured: relative ordering and blockwise response magnitude capture different aspects of a model's development. Cross-domain transfer further identifies a shared component of redistribution: changes measured on two text domains correlate with the relative changes on a third at the same training endpoints.

This longitudinal contrast raises a second question: how should a changing bypass effect be interpreted? Removing a block introduces a missing-update perturbation into the residual stream, which the remaining layers propagate to the output. We study these components in six Pythia-160M training runs and a contrasting OLMo-2 7B trajectory. Continuous output metrics connect the depth-profile observations to changes in the output distribution. In Pythia, matching relative perturbation magnitude reveals that an increasing bypass effect accompanies a smaller pooled downstream response, with the contrast depending on perturbation strength. The intervention pathway therefore matters when interpreting sensitivity changes over training.

We make three contributions. First, across five released pretraining trajectories and 11 model--domain combinations, we characterize retained depth ordering alongside concentrated changes in blockwise response. Second, we show that large-change positions recur on held-out text windows and that redistribution patterns transfer across evaluation domains within a trajectory. Third, controlled perturbations distinguish changes in local missing-update magnitude from matched downstream response. Their opposing pooled trends in Pythia and scale dependence establish why bypass effects should be interpreted jointly with the perturbations that produce them.

These controlled tests leave the causes of the broad cross-domain pattern open. The measured responses do not identify semantic circuits, causal shares, targeted compensation, or safe joint pruning.

\section{Related work}
\paragraph{Depth-dependent interventions.}
Layer deletion, swapping, and similarity-based removal reveal nonuniform effects across Transformer depth \citep{gromov2025deeper,men2025shortgpt,lad2025remarkable}. Sublayer attribution provides a finer-grained view of this variation \citep{zhang2024}. Divergent Token Metrics evaluates output fidelity using intact-generated histories \citep{deiseroth2024dtm}, while mathematical ablations compare task-critical layers across base and post-training models \citep{nepal2025layer}. These studies establish nonuniform depth sensitivity. We ask how the ordering and magnitudes of complete block-bypass profiles develop along pretraining trajectories.

\paragraph{Longitudinal interpretability.}
Induction heads, grokking, and circuit analyses connect internal computations to learning over time \citep{olsson2022,nanda2023progress,tigges2024circuits}. Sparse crosscoders align learned features across checkpoints \citep{bayazit2026}, and layer-dropping studies track redundancy during training \citep{he2026}. At the whole-layer level, \citet{garcia2026nofreeswap} compare replacement and interchange protocols across checkpoints; \citet{spiro2026functional} study Pythia ablations in a public author manuscript. We follow structurally aligned block positions to ask which aspects of a complete response profile persist and where its changes concentrate across evaluation domains.

\paragraph{Intervention metrics and downstream response.}
Activation-patching conclusions depend on the intervention and output metric \citep{zhang2023patching,makelov2024subspace}. Self-repair studies show how downstream changes modify a component's total ablation effect \citep{rushing2024selfrepair}. We measure the missing update at a block boundary and the suffix response at a matched relative perturbation magnitude. These interventions distinguish components of the measured response pathway. They connect the longitudinal profile to local update magnitude and downstream response, providing experimentally controlled views of how block sensitivity develops.

\section{Methods and experimental setup}
\label{sec:methods}
\subsection{Block bypass and sensitivity profiles}
At checkpoint $t$, block $\ell$ maps its residual-stream input $h_\ell$ to $h_\ell^+$. We compare the intact model with one that replaces this block by the identity:
\begin{equation}
 h_\ell^+=B_\ell^{(t)}(h_\ell),\qquad h_{\ell,\mathrm{bypass}}^+=h_\ell.
 \label{eq:bypass}
\end{equation}
Each forward pass bypasses one complete block without retraining. For an architecture with $H$ blocks, we evaluate the $L=H-2$ interior positions, indexed as $\mathcal I=\{1,\ldots,H-2\}$. Architecture, tokenizer, block positions, and tokenized contexts remain fixed within each trajectory and text domain.

Let $p_i^{(t)}$ and $q_{i,\ell}^{(t)}$ denote the intact and bypassed next-token distributions at prediction position $i$. We measure top-1 disagreement,
\begin{equation}
 D_\ell(t)=\frac1N\sum_{i=1}^N
 \ind\!\left[\arg\max_v p_i^{(t)}(v)\ne\arg\max_v q_{i,\ell}^{(t)}(v)\right],
 \quad \vectD(t)=(D_\ell(t))_{\ell\in\mathcal I}.
 \label{eq:D}
\end{equation}
Teacher forcing gives both models the same history at every position. Each profile uses 18 fixed windows of 257 tokens, yielding $N=4{,}608$ next-token predictions per block. The early overview in Figure~\ref{fig:workflow} summarizes the measurement and paper roadmap.

\subsection{Models and evaluation text}
The broad analysis covers five released trajectories from OLMo-2, SmolLM2, and Pythia \citep{olmo2025,olmo1b2025,smol2025,biderman2023pythia}. Five checkpoints span approximately 10\%, 25\%, 50\%, 75\%, and 100\% of the evaluated training stage; analyses retain the recorded progress coordinates. General text comes from WikiText-2, mathematical text from GSM8K question--answer examples, and code text from MBPP text--code examples \citep{merity2017wikitext,cobbe2021gsm8k,austin2021mbpp}. These are next-token evaluations on each text domain. In total, the 11 model--domain combinations provide 55 profiles and 1,290 block-level measurements (Table~\ref{tab:coverage}).

\begin{table}[!htb]
\caption{\textbf{Model and text coverage.} G/M/C denote general, mathematical, and code text. The released models each contribute one training trajectory. Pythia-160M provides repeated training runs for the controlled experiments.}
\label{tab:coverage}
\centering\small
\begin{tabular}{@{}lcccc@{}}
\toprule
Model & Measured blocks & Checkpoints & Text & Runs \\
\midrule
OLMo-2 1B & 14 & 5 & G/M/C & 1 \\
OLMo-2 7B & 30 & 5 & G/M/C & 1 \\
OLMo-2 13B & 38 & 5 & G & 1 \\
SmolLM2 1.7B & 22 & 5 & G/M/C & 1 \\
Pythia 1.4B & 22 & 5 & G & 1 \\
\midrule
Pythia 160M & 10 & 3 & Hella-derived & 6 \\
Pythia 160M & 10 & 3 & WikiText-2 & 5\textsuperscript{*} \\
\bottomrule
\end{tabular}
\par\smallskip\footnotesize\textsuperscript{*}Two runs also occur in the Hella-derived evaluation.
\end{table}

The replicated Pythia-160M evaluations use steps 14k, 72k, and 143k on Hella-derived text \citep{zellers2019hellaswag} and corrected WikiText-2 measurements. Continuous-output and controlled-perturbation experiments use the six Hella run pairs at 14k and 143k, together with OLMo-2 7B at 391B and 3896B training tokens. Appendix~\ref{app:data} provides checkpoint identities, text construction, and execution precision.

\subsection{Comparing profiles over training}
We treat the measurements as a checkpoint-by-depth response map and characterize its overall level, depth ordering, and localized change. We measure overall level, $\bar D(t)=L^{-1}\sum_\ell D_\ell(t)$, and depth ordering using Spearman correlation. Correlations are grouped by checkpoint-index separation. This comparison asks how well the ordering persists across increasingly distant observations.

The endpoint change $\delta_\ell=D_\ell(T)-D_\ell(1)$ measures response redistribution across positions, without assuming that total response is conserved. We calculate the share of absolute change carried by the largest-changing $k=\lceil0.2L\rceil$ positions. To test whether these positions recur on different text windows, we select them on nine windows and evaluate their share on the other nine. Repeating 1,000 random partitions measures the consistency of this held-out concentration.

For cross-domain transfer, $z(v)$ standardizes a vector across depth. A template formed from two domains is compared with the third domain's endpoint change:
\begin{equation}
 F^{(-d)}=\tfrac12\!\sum_{d'\ne d}z(\delta^{(d')}),
 \qquad r_d=\operatorname{Pearson}(F^{(-d)},\delta^{(d)}).
 \label{eq:domain}
\end{equation}
This tests agreement in the location and relative magnitude of change. Each comparison holds the model and checkpoint pair fixed.

The broad profile analyses are retrospective summaries of recorded measurements. Uncertainty follows the source of replication. For six model--domain combinations with complete window records, 2,000 paired window-bootstrap draws retain alignment across layers and checkpoints. These window intervals describe within-trajectory text variation. For the controlled Pythia experiments, intervals resample the six paired training runs. OLMo remains a single-trajectory comparison. Trajectory-balanced summaries and secondary profile analyses appear in Appendix~\ref{app:profiles}.

\subsection{Continuous outputs and matched perturbations}
We measure continuous output change with Jensen--Shannon divergence,
\begin{equation}
 \JS(p,q)=\tfrac12\operatorname{KL}(p\|m)+\tfrac12\operatorname{KL}(q\|m),
 \qquad m=\tfrac12(p+q).
 \label{eq:js}
\end{equation}
Forward KL, target-token negative-log-likelihood change, and loss of the intact top-versus-runner-up logit margin provide additional checks. Metrics are computed in FP32. Pythia meets the revised representation-derived numerical reproducibility criterion (Appendix~\ref{app:continuous}); OLMo recovers historical counts exactly. Five intact-margin strata use the same early-checkpoint quantile cutpoints at both endpoints, allowing comparisons at similar initial prediction confidence.

A natural bypass introduces $\delta_{\rm nat}=h_\ell-h_\ell^+$ at the block output. We measure its root-mean-square (RMS) magnitude relative to the intact residual state and define a shared target for each paired token and block:
\begin{equation}
 r_{\rm nat}=\frac{\RMS(h_\ell-h_\ell^+)}{\RMS(h_\ell^+)+10^{-8}},\qquad
 r^*(\gamma)=\gamma\min(r_{\rm nat}^{\rm early},r_{\rm nat}^{\rm final}),
 \label{eq:match}
\end{equation}
where $\gamma\in\{0.25,0.50,0.75,1\}$. Each checkpoint's natural direction is rescaled to this common relative magnitude. We measure the perturbation after FP16 rounding and adjust JS for its actual relative RMS. The primary comparison averages over all four scales. Relative RMS normalizes the missing update by the intact residual state. Matching asks how much output change follows a common relative perturbation while retaining each checkpoint's own state and natural direction. Appendices~\ref{app:continuous}--\ref{app:directions} detail numerical validation, matching, and directional comparisons.

\section{Results}
\subsection{Depth ordering persists across changing response levels}
\label{sec:persistence}
\begin{figure}[!htb]
\centering\threeplots{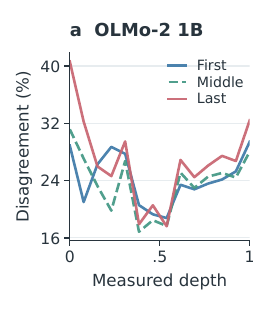}{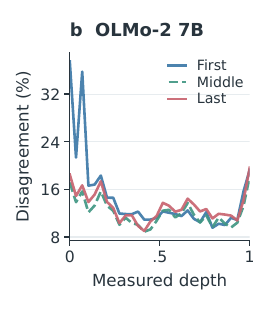}{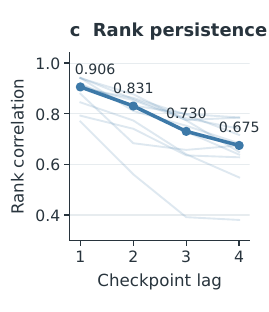}
\caption{\textbf{Depth ordering persists as response magnitudes evolve.} (a,b) General-text profiles at the first, middle, and last observed checkpoints. Depth is normalized over the measured interior blocks. (c) Thin lines show each model--domain combination's median correlation at a given checkpoint separation; the heavy line summarizes all pairs. Separations 1--4 contain 44, 33, 22, and 11 pairs.}
\label{fig:persistence}
\end{figure}

The relative ordering of block sensitivities remains positively associated across the observed training stages (Figure~\ref{fig:persistence}). Across the 11 released model--domain combinations, median Spearman correlation is 0.906 for adjacent checkpoints and 0.831, 0.730, and 0.675 at separations of two, three, and four checkpoints. Every endpoint correlation is positive, ranging from 0.380 to 0.785. Within every combination, mean adjacent correlation exceeds endpoint correlation. Nearby checkpoints thus share more of the depth ordering than the observed endpoints. This relationship summarizes persistence across the sampled training stages.

Ordering persists under both increasing and decreasing overall sensitivity. Mean disagreement increases in eight combinations and decreases in all three OLMo-2 7B domains. On general text, OLMo-2 7B changes from 14.57\% to 13.11\%, whereas OLMo-2 1B changes from 24.28\% to 26.65\%. Giving each trajectory equal weight yields median adjacent and endpoint correlations of 0.894 and 0.649. Paired-window uncertainty and all endpoint summaries appear in Appendix~\ref{app:profiles}. Rank correlation tracks the ordering of the same positions, so retained rankings can coexist with changes in response level and spacing. We next examine the magnitude and location of these changes.

\subsection{Localized redistribution recurs across windows and domains}
\label{sec:redistribution}
\begin{figure}[!htb]
\centering\twoplots{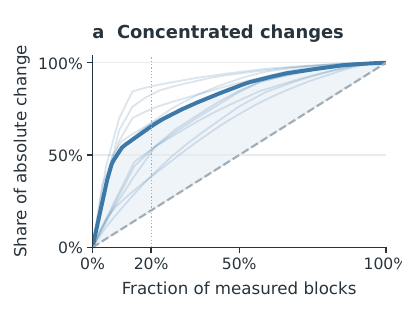}{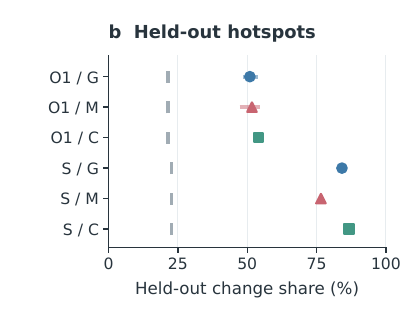}
\caption{\textbf{Response changes concentrate at repeatable positions.} (a) Cumulative absolute endpoint change, with blocks ordered by change magnitude. Thin curves show model--domain combinations; the dark curve is their pointwise median. The diagonal represents equal change at every block. (b) The largest-changing positions are selected on nine windows and evaluated on the other nine. Points and bars show medians and 5th--95th percentiles over 1,000 partitions. Gray ticks show the random-selection expectation. O1/S denote OLMo-2 1B/SmolLM2; G/M/C denote the three text domains.}
\label{fig:concentration}
\end{figure}

Individual block responses change substantially within the persistent ordering. The largest-changing $\lceil0.2L\rceil$ positions account for a median 65.5\% of absolute endpoint change (Figure~\ref{fig:concentration}a). A global shift and rescaling of the first profile leaves substantial depth-specific residuals; the corresponding fits are shown in Appendix~\ref{app:profiles}. Training therefore changes the numerical separation between block responses as well as their overall level.

The same positions also carry large changes on held-out windows. Their held-out shares range from 51.0\% to 54.0\% across the OLMo-2 1B domains and from 76.6\% to 86.8\% across SmolLM2 (Figure~\ref{fig:concentration}b). Random sets of the same sizes have expected shares of 21.4\% and 22.7\%, respectively. These held-out shares support reproducible positions of change within the available text samples. Changes frequently concentrate toward the front of the measured depth range: the shallow third carries the largest share in nine of 11 combinations. Regional differences and boundary-removal controls are reported in Appendix~\ref{app:profiles}.

\begin{figure}[!htb]
\centering\threeplots{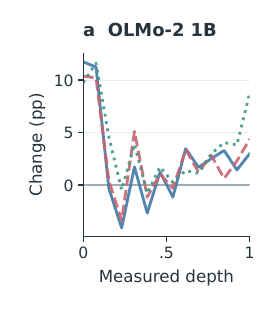}{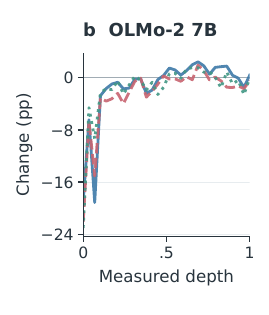}{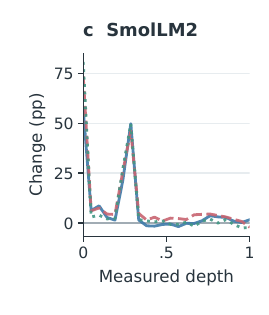}
\par\includegraphics[width=\linewidth]{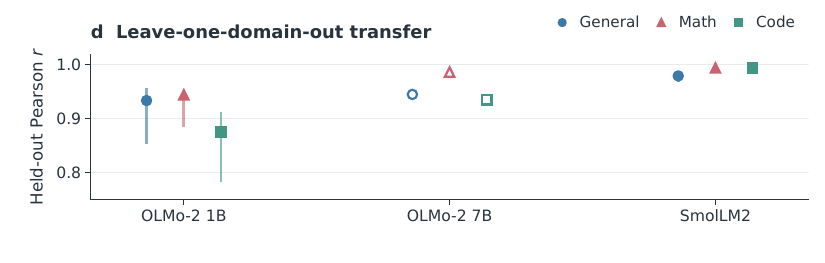}
\caption{\textbf{A trajectory's redistribution pattern transfers across text domains.} (a--c) Endpoint changes in percentage points, with model-specific vertical scales. Blue, red, and green denote general, mathematical, and code text. (d) Each domain is correlated with a standardized template constructed from the other two. Filled points include 95\% paired-window-bootstrap intervals; hollow points show estimates with unavailable window-level uncertainty. The template and held-out domain use the same training endpoints.}
\label{fig:domains}
\end{figure}

\paragraph{Transfer across evaluation domains.}
\label{sec:signature}
The three trajectories evaluated on general, mathematical, and code text exhibit closely aligned endpoint changes (Figure~\ref{fig:domains}). Across nine within-trajectory domain pairs, the change vectors have median Pearson correlation 0.954. OLMo-2 7B loses sensitivity most strongly near the front of the measured range. SmolLM2 instead develops large increases at a few shared positions. The commonality is therefore in \emph{where and in which direction} sensitivity changes, even when domains differ in overall response.

A template constructed from two domains correlates with the held-out third domain at a median 0.944, with all nine comparisons between 0.875 and 0.993. Removing the first and last measured blocks retains strong transfer (median 0.959). Thus the agreement extends beyond a single boundary response. Comparisons of the four successive checkpoint changes also show shared depth patterns across domains (Appendix~\ref{app:fields}). Together, these results identify a shared component of redistribution within each trajectory: depthwise changes measured on two text domains predict the pattern on a held-out domain.

Subtracting each domain's initial profile removes its time-constant depth baseline. The standardized template then compares relative gains and losses across blocks. High held-out correlation means that positions with comparatively large changes in two domains also tend to change strongly in the third, linking the result directly to redistribution over training.

\subsection{Continuous outputs retain both the training effect and depth ordering}
\label{sec:continuous}
\begin{figure}[!htb]
\centering\threeplots{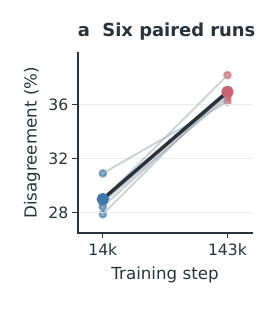}{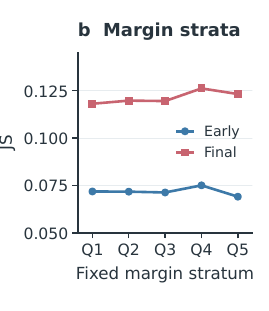}{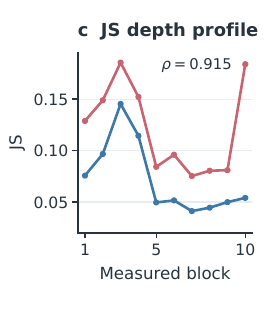}
\caption{\textbf{The training effect extends to continuous output distributions.} (a) Pythia disagreement in six paired runs, with their mean shown by the heavy line. (b) JS increases in every intact-margin stratum, using fixed early-checkpoint cutpoints. (c) Run-averaged JS profiles retain depth ordering while increasing in magnitude ($\rho=0.915$). Blue and red denote early and final checkpoints.}
\label{fig:continuous}
\end{figure}

We next test whether longitudinal sensitivity is reflected in continuous output distributions. In the continuous-output evaluation using the revised reproducibility criterion, mean disagreement across six Pythia-160M training runs increases from 28.99\% at step 14k to 36.91\% at step 143k. The paired change is 7.92 percentage points, with a 95\% run-bootstrap interval of [6.77, 8.81], and is positive in every run (Figure~\ref{fig:continuous}a). Separate historical three-checkpoint Hella and corrected WikiText-2 evaluations show the same direction (Appendix~\ref{app:data}).

Mean JS increases from 0.0723 to 0.1216, a change of 0.0493 with a 95\% paired-run interval of [0.0435, 0.0537]. Forward KL, target-token NLL change, and intact-competitor margin loss also increase in all six runs. Both disagreement and JS increase within every fixed intact-margin stratum (Figure~\ref{fig:continuous}b). The training effect is consequently present at comparable intact margins and in the full output distribution. Agreement across these metrics shows that the change extends beyond top-token flips to the intact model's probability assignments on the evaluated contexts.

Continuous depth profiles preserve their ordering as well. Pythia's run-averaged JS profiles have endpoint rank correlation 0.915, alongside 0.891 for forward KL and 0.879 for target NLL change. OLMo-2 7B decreases across all five output measures while retaining associated JS depth ranks ($\rho=0.766$). These two settings show that retained ordering can accompany either direction of continuous response change. Complete metric comparisons are in Appendix~\ref{app:continuous}.

\subsection{Local perturbations and downstream responses change differently}
\label{sec:decomposition}
\begin{figure}[!htb]
\centering\threeplots{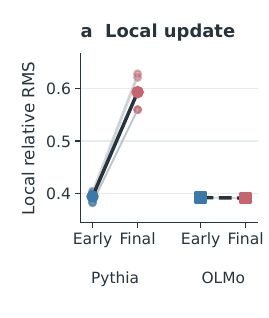}{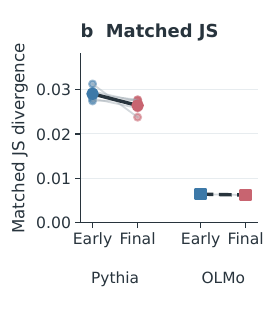}{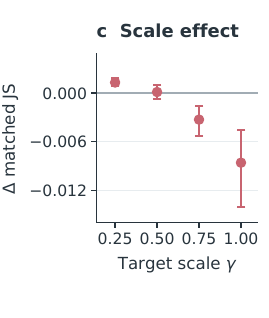}
\caption{\textbf{Training changes the balance between local perturbation and downstream response.} (a) Natural local relative RMS. (b) JS at matched relative magnitude, adjusted for the perturbation after FP16 rounding and pooled over four scales. Thin Pythia lines show runs; heavy lines show their means. OLMo contributes one trajectory. (c) Pythia checkpoint changes at each target scale, with paired-run intervals.}
\label{fig:decomposition}
\end{figure}

The representation-aware matched-perturbation experiment separates changes in the missing update from changes in downstream response (Figure~\ref{fig:decomposition}). Pythia's natural local relative RMS grows from 0.3946 to 0.5927. The increase of 0.1982 has a 95\% paired-run interval of [0.1827, 0.2140] and occurs in every run. At matched relative magnitude, the pooled JS response decreases from 0.02903 to 0.02641. Its adjusted change is $-0.00262$, with interval [$-0.00478,-0.00100$], and is negative in five of six runs. Thus a growing natural output response accompanies a larger missing update and a smaller pooled matched response.

OLMo-2 7B exhibits a different balance. Local relative RMS remains nearly unchanged (0.3925 to 0.3915), while matched JS decreases from 0.006424 to 0.006200 and natural JS decreases from 0.02329 to 0.01799. The opposite natural-response trends are therefore accompanied by different local changes within the same intervention design.

Perturbation strength matters. For Pythia, the matched-JS change is positive at $\gamma=0.25$ (+0.00129), its interval includes zero at $0.50$, and it is negative at $0.75$ and $1$ ($-0.00329$ and $-0.00861$). The pooled decrease summarizes this four-scale response. A model's downstream sensitivity can consequently change differently for small and larger relative perturbations along the natural missing-update direction.

These experiments show why the natural bypass effect and matched response provide complementary information. In Pythia, larger missing updates accompany a growing natural effect and a smaller pooled matched response; the scale-resolved results locate where this contrast occurs. Direction controls show depth- and control-family-dependent attenuation. Neither checkpoint meets the preregistered joint criterion for targeted natural-direction compensation. This supports a scale- and direction-dependent downstream description, without identified causal shares or self-repair. Propagation analyses and the plotted direction controls appear in Appendices~\ref{app:matching} and~\ref{app:directions}.

\section{Discussion}
\label{sec:discussion}
\paragraph{Persistence is a property of the measured organization.}
Across the released trajectories, layer rankings remain associated while response magnitudes change at positions that recur across evaluated text. Cross-domain transfer adds a temporal dimension: different texts reveal similar changes in the profile. These results distinguish where a model tends to respond from where that response changes. An enduring ordering can coexist with changes in individual responses and selected block identities.

\paragraph{The intervention pathway also develops during training.}
A bypass removes a learned update, and its output effect depends on the remaining network. Pythia exhibits strong local perturbation growth alongside a decreasing pooled matched response; OLMo exhibits nearly stable local magnitude and decreasing matched and natural responses. The scale-dependent sign change in Pythia shows why perturbation strength belongs in this description. Matching relative magnitude controls one aspect of the intervention while retaining checkpoint-specific directions and residual states. The results support a local-plus-suffix account, not an identified allocation of causal shares.

These findings complement circuit-level observations of persistent computations under changing implementations \citep{tigges2024circuits}. Here, shared redistribution provides a measured pattern for following depth organization across checkpoints. The controlled tests show how different components evolve within the intervention pathway; they leave the causes of the broad cross-domain pattern open. Direction controls show depth- and control-family-dependent propagation without identifying targeted downstream compensation.

\paragraph{Scope and limitations.}
The broad comparisons cover single released trajectories; independent-run replication is concentrated in Pythia-160M, and the two text panels share run identities. Window repeatability does not establish that larger-model change locations recur across training seeds. Families differ jointly in architecture, data, schedule, tokenizer and precision. The experiments measure next-token response on fixed teacher-forced text, rather than mathematical/code correctness or free-running generation. Whole-block bypass does not isolate semantic circuits, and separate bypass costs do not establish safe joint pruning. Checkpoints and text samples are finite; matched interventions retain each checkpoint's state and direction, and the tested controls do not exhaust possible geometry.

\section{Conclusion}
Across the observed pretraining stages, block-bypass profiles retain associated depth ordering while their numerical responses undergo concentrated changes shared across evaluation domains. Continuous-output measurements reproduce the coexistence of retained ordering and changing response magnitude in the two focal settings. Matching perturbation magnitude reveals opposing pooled local and downstream changes in Pythia, with a scale-dependent contrast and a different balance in OLMo. These findings connect longitudinal depth organization to an evolving intervention pathway without identifying targeted compensation or a universal increase in layer dependence.

\label{maintextend}
\clearpage
\section*{Reproducibility statement}
Section~\ref{sec:methods} specifies the intervention, evaluation data, profile comparisons, and controlled perturbation design. Appendices~\ref{app:data}--\ref{app:directions} give checkpoint identities, aggregation, numerical validation, matching rules, and directional comparisons. The accompanying package contains the complete released-profile table, the 35 exported window-count arrays, focal preregistrations and processed summaries, figure and table code, numerical checks, editable manuscript source, and vector figures with plotting inputs. The provided data reproduce the plotted broad-profile analyses and focal summary displays. Full focal acquisition and token-level model replay require the retained raw experiment records and acquisition code, which are outside this summary export.

\section*{AI use statement}
Generative AI tools assisted with developing the conceptual framing, proposing and refining hypotheses, reviewing experimental methodology, implementing and checking analysis code, organizing numerical data, interpreting results, literature verification, translation, and manuscript and figure preparation. The empirical measurements used here originate from the recorded model experiments; no synthetic experimental outcomes were generated for the manuscript. The package provides source-linked numerical checks and executable analysis and plotting code for the included summaries. The human authors retain responsibility for the final scientific claims, references, disclosures, and submitted artifacts.

\section*{Ethics statement}
This study analyzes language-model interventions using existing benchmark text and recorded model checkpoints. The released numerical export omits underlying benchmark text and private execution provenance. Applications involving model modification should evaluate the relevant capabilities and safety properties directly; bypass-response summaries alone do not provide that assessment.

{\small\raggedright
\bibliographystyle{iclr2027_conference}
\bibliography{references}}
\clearpage
\appendix
\section{Evaluation data and reproducibility}
\label{app:data}
\begin{figure}[p]
\centering\includegraphics[width=\linewidth]{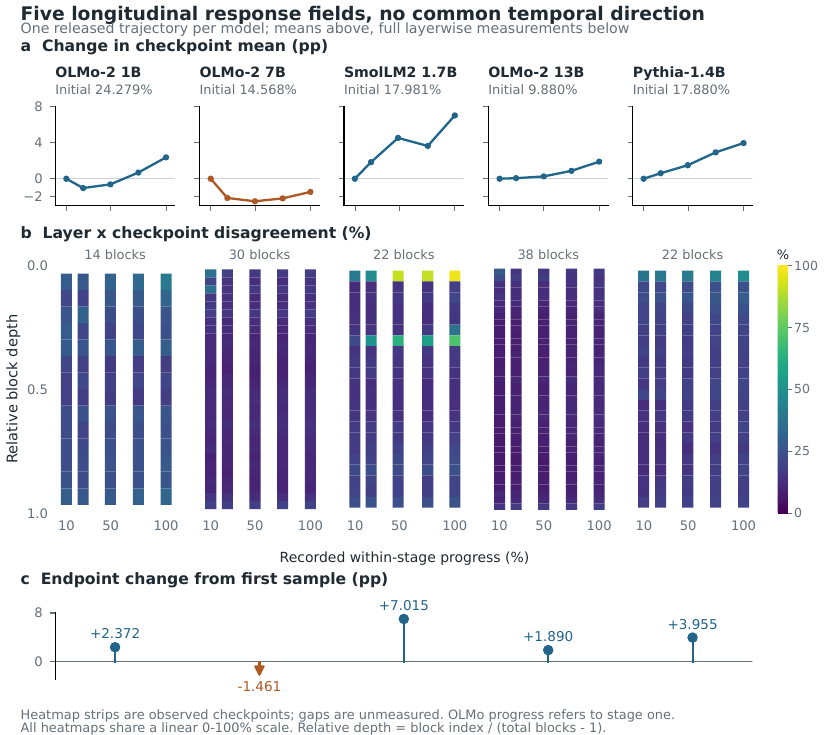}
\caption{\textbf{Five released general-text response fields.} (a) Mean changes share vertical limits and use recorded within-stage progress; initial absolute means are printed above each model. (b) All 630 interior-block/checkpoint responses use a common 0--100\% scale. Relative depth is block index divided by total blocks minus one. Narrow strips denote the five observed samples; gaps are unmeasured. (c) Endpoint stems are descriptive point estimates, not confidence intervals. OLMo-2 7B ends lower despite partial recovery; OLMo progress refers to stage one. Each model supplies one trajectory.}
\label{fig:dense-released}
\end{figure}

\paragraph{Released trajectories.}
The five released trajectories are listed in Table~\ref{tab:coverage}; complete general-text response fields appear in Figure~\ref{fig:dense-released}. We exclude the first and last architectural blocks and include every interior position. The first and last observed checkpoints delimit the evaluated release or training stage. OLMo-2 and SmolLM2 use FP16 inference. Pythia-1.4B uses FP32 throughout after an initial FP16 evaluation produced nonfinite logits. Execution precision is fixed within each trajectory.

The supplementary data include repository and checkpoint identifiers, recorded training progress, evaluated block positions, precision, context hashes, and checksums. The complete profile table contains 1,290 measurements. This released export contains window-level disagreement counts for 35 profiles, with 18 windows and 256 prediction positions per window. OLMo-2 1B and SmolLM2 have complete window records at all five checkpoints in all three domains. Four OLMo-2 13B checkpoints and one Pythia-1.4B checkpoint have partial coverage; the OLMo-2 7B data in this export include profile summaries. Window-resampling analyses consequently use the six complete model--domain combinations.

\paragraph{Training runs and text construction.}
Pythia-160M training runs are identified by initialization and training-data-order seeds. The six Hella-derived evaluations use runs 1, 4, 6, 7, 8, and 9. Runs 1, 4, and 9 originated in pilot experiments; runs 6, 7, and 8 were confirmatory. The reported six-run summaries include both groups. Corrected WikiText-2 evaluations use runs 2, 3, 5, 6, and 8, sharing runs 6 and 8 with the Hella-derived evaluations. Three-checkpoint profiles cover steps 14k, 72k, and 143k. The continuous-output and matched-perturbation experiments compare 14k with 143k on the same six Hella run pairs.

Hella-derived text concatenates validation contexts with labeled correct endings, so windows may cross example boundaries. The 18 evaluation windows use text-sampling seeds 11, 23, and 37, with six windows per seed. The continuous-output, norm-matching, and direction-control experiments share these windows. The OLMo-2 7B comparison uses checkpoints at 391B and 3896B tokens, with exact revision identifiers in the supplement. Reusing runs and contexts allows paired comparisons across interventions.

\paragraph{Repeated evaluations and numerical reproducibility.}
Figure~\ref{fig:continuous} uses the continuous-output experiment described in Section~\ref{sec:continuous}. Its disagreement change is 7.92173 percentage points, with interval [6.77083, 8.80896], and its depth-rank correlation is 0.90303. The earlier three-checkpoint Hella evaluation has change 7.92643, interval [6.76722, 8.85091], and rank correlation 0.86667. These estimates come from separately recorded executions. The earlier interval uses 100,000 paired-run bootstrap samples with seed 20260828. Corrected WikiText-2 disagreement increases from 34.29\% to 43.58\% in all five runs.

The corrected WikiText-2 evaluation uses the same lowest-index argmax rule on intact and bypassed paths. Measurements from an earlier implementation with inconsistent tie handling are excluded. Historical Hella token spans and hashes are retained. Corrected WikiText-2 can be reconstructed from the retained inputs and code, but a separate token-window manifest and complete environment lockfile are unavailable. Appendix~\ref{app:continuous} describes the numerical checks for the continuous-output experiment.

\paragraph{Supplementary material.}
The package includes released-profile data, 35 exported window-count arrays, analysis code, empirical vector figures and their plotting inputs. The former HTML/SVG schematics are not used. The controlled experiments are documented by their preregistered protocols, numerical checks, processed summaries, and checkpoint references. Their raw token-level activations and acquisition code remain in the retained experiment archive. The supplementary README maps protocol versions to the results used in this paper.

\section{Additional profile analyses}
\label{app:profiles}
\begin{figure}[!htb]
\centering\twoplots{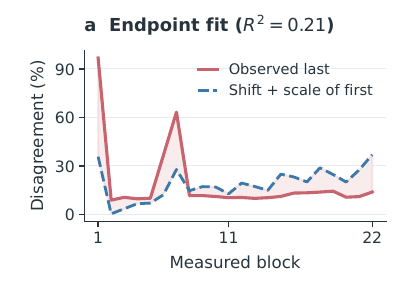}{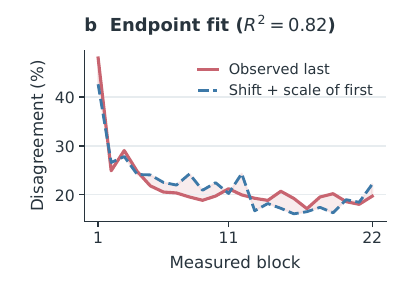}
\caption{\textbf{A common shift and scale leave depth-specific residuals.} SmolLM2/code and Pythia-1.4B/general illustrate the lowest and highest affine $R^2$ values. Red shows the observed final profile; blue shows the fitted transformation of the first profile. Shading marks their difference.}
\label{fig:affine}
\end{figure}

\paragraph{Overall response and affine correspondence.}
Table~\ref{tab:broadsummary} reports the endpoint summaries for every model--domain combination. To quantify the change left after a common shift and scale, we fit
\begin{equation}
 D_\ell(T)=a+bD_\ell(1)+e_\ell
\end{equation}
across blocks. The reported $R^2$ is the fraction of final between-block variance explained by this fit. It has median 0.461 and range 0.206--0.821. Figure~\ref{fig:affine} shows the two ends of that range. Substantial residuals identify depth-specific changes even when the profiles retain a positive rank correlation.

\paragraph{Concentration of change.}
For $\delta_\ell=D_\ell(T)-D_\ell(1)$ and the $k=\lceil0.2L\rceil$ largest-changing positions $K$, we calculate
\begin{equation}
 C_k=\frac{\sum_{\ell\in K}|\delta_\ell|}{\sum_{\ell\in\mathcal I}|\delta_\ell|}.
 \label{eq:concentration}
\end{equation}
The selected fractions are $3/14$, $6/30$, $8/38$, or $5/22$, depending on architecture. Selection by observed change concentrates the same-sample statistic by construction. The held-out-window analysis in Figure~\ref{fig:concentration} tests whether those selected locations carry large changes on other windows. A uniformly random set of $k$ positions has expected share $k/L$ for any fixed held-out change vector.

\begin{table}[!htb]
\caption{\textbf{Endpoint response, redistribution, and identity retention.} High/Low are the overlap fractions of the highest/lowest response sets. G/M/C denote general, mathematical, and code text. The affine fit measures numerical correspondence after a common shift and scale.}
\label{tab:broadsummary}
\centering\small\begin{tabular}{@{}lrrrrrr@{}}
\toprule
Panel & $\Delta\bar D$ (pp) & $\rho$ & Affine $R^2$ & $C_k$ (\%) & High & Low \\
\midrule
OLMo-2 1B / G & +2.37 & 0.675 & 0.461 & 54.5 & 0.67 & 1.00 \\
OLMo-2 1B / C & +3.77 & 0.741 & 0.570 & 54.6 & 1.00 & 0.67 \\
OLMo-2 1B / M & +2.65 & 0.380 & 0.411 & 54.9 & 0.33 & 0.33 \\
OLMo-2 13B / G & +1.89 & 0.716 & 0.622 & 39.6 & 0.75 & 0.38 \\
OLMo-2 7B / G & -1.46 & 0.785 & 0.543 & 67.3 & 1.00 & 0.50 \\
OLMo-2 7B / C & -1.71 & 0.628 & 0.303 & 65.5 & 0.83 & 0.50 \\
OLMo-2 7B / M & -2.47 & 0.785 & 0.397 & 67.3 & 0.83 & 0.83 \\
SmolLM2 / G & +7.02 & 0.639 & 0.687 & 84.9 & 0.80 & 0.20 \\
SmolLM2 / C & +7.66 & 0.683 & 0.206 & 88.1 & 0.60 & 0.60 \\
SmolLM2 / M & +9.25 & 0.548 & 0.395 & 76.6 & 0.60 & 0.60 \\
Pythia 1.4B / G & +3.95 & 0.649 & 0.821 & 42.9 & 0.60 & 0.20 \\
\bottomrule
\end{tabular}

\end{table}

\paragraph{Paired-window uncertainty.}
Each bootstrap sample draws 18 windows with replacement and uses the same multiplicities for every block and checkpoint within a domain. Different domains are sampled independently. The 2,000 draws use seed 20260925. Table~\ref{tab:windowvalidation} gives percentile 95\% intervals for the difference between mean adjacent-checkpoint correlation and endpoint correlation. All six complete model--domain combinations have positive intervals, supporting greater correspondence over shorter sampled intervals.

The held-out concentration analysis divides the 18 windows into two disjoint halves. It selects the $k$ largest-changing positions on one half, measures their absolute-change share on the other, and averages the two directions. We report the median and 5th--95th percentiles over 1,000 partitions. These describe sensitivity to the partition of the available text windows. To prevent trajectories with three domains from receiving greater weight, we also average domain summaries within each trajectory before taking the median across trajectories. This gives adjacent and endpoint correlations of 0.894 and 0.649, and same-sample change concentration of 54.7\%.

\begin{table}[!htb]
\caption{\textbf{Window-level evidence for persistence and localized change.} Correlation-gap intervals use paired-window bootstrapping. The remaining columns show held-out change shares and their partition percentiles.}
\label{tab:windowvalidation}
\centering\small\begin{tabular}{@{}lrrrr@{}}
\toprule
Panel & $\bar\rho_{\rm adj}-\rho_{\rm end}$ & 95\% interval & Share (\%) & 5--95\% splits \\
\midrule
OLMo-2 1B / G & 0.171 & [0.091, 0.323] & 51.0 & [48.6, 53.9] \\
OLMo-2 1B / C & 0.201 & [0.107, 0.305] & 54.0 & [52.9, 54.7] \\
OLMo-2 1B / M & 0.402 & [0.176, 0.423] & 51.7 & [47.2, 54.7] \\
SmolLM2 / G & 0.283 & [0.202, 0.384] & 84.2 & [83.4, 84.7] \\
SmolLM2 / C & 0.231 & [0.172, 0.315] & 86.8 & [85.3, 87.5] \\
SmolLM2 / M & 0.298 & [0.207, 0.402] & 76.6 & [75.6, 76.8] \\
\bottomrule
\end{tabular}

\end{table}

\paragraph{Location of change.}
The shallow region ends at $\lceil L/3\rceil$ and the middle region at $\lceil2L/3\rceil$. Regional shares sum absolute endpoint changes over these positions. In the boundary-removal check, the first measured block is removed from both the regional numerator and total denominator, keeping the regional boundaries fixed. The shallow third carries the largest share in nine combinations before and after removal. OLMo-2 13B and Pythia-1.4B have more middle or deep change (Table~\ref{tab:regions}). The evidence therefore supports concentrated redistribution with a frequent shallow concentration, alongside trajectory-specific spatial patterns.

\begin{table}[!htb]
\caption{\textbf{Where endpoint changes accumulate.} The trimmed column gives the shallow share after excluding the first measured interior block. Regional boundaries are unchanged.}
\label{tab:regions}
\centering\small\begin{tabular}{@{}lrrrr@{}}
\toprule
Panel & Shallow (\%) & Middle (\%) & Deep (\%) & Shallow, trimmed (\%) \\
\midrule
OLMo-2 1B / G & 58.7 & 20.6 & 20.7 & 45.8 \\
OLMo-2 1B / C & 54.8 & 9.6 & 35.5 & 45.2 \\
OLMo-2 1B / M & 63.1 & 15.2 & 21.7 & 52.5 \\
OLMo-2 13B / G & 31.3 & 41.4 & 27.4 & 24.0 \\
OLMo-2 7B / G & 70.0 & 14.4 & 15.5 & 60.3 \\
OLMo-2 7B / C & 69.2 & 16.5 & 14.3 & 53.5 \\
OLMo-2 7B / M & 77.6 & 9.7 & 12.7 & 69.2 \\
SmolLM2 / G & 88.2 & 4.1 & 7.7 & 82.1 \\
SmolLM2 / C & 91.3 & 3.2 & 5.5 & 84.6 \\
SmolLM2 / M & 83.0 & 7.4 & 9.6 & 73.6 \\
Pythia 1.4B / G & 32.3 & 29.9 & 37.8 & 24.0 \\
\bottomrule
\end{tabular}

\end{table}

\paragraph{Selected block identities.}
The highest- and lowest-response sets each contain $k$ blocks, with ties resolved by block index. Retention is the intersection of early and final sets divided by $k$. Median high-set retention is 0.75 and low-set retention is 0.50. Because near-tied block responses are sensitive to the text sample, we also compare memberships at a matched nine-window budget (Table~\ref{tab:identity}). Within-checkpoint overlap compares the two halves at the same checkpoint and averages early and final values. Across-checkpoint overlap compares one half early with the other half final and averages the two directions. This separates the observed temporal change in membership from variation due to the finite evaluation sample.

\begin{table}[!htb]
\caption{\textbf{Block-set stability with matched text budgets.} Entries are medians over 1,000 half-window partitions. High/Low use the same set sizes as Table~\ref{tab:broadsummary}. Within-checkpoint overlap provides a text-sampling comparison for across-checkpoint retention.}
\label{tab:identity}
\centering\small\begin{tabular}{@{}lrrrr@{}}
\toprule
Panel & High: within & High: across & Low: within & Low: across \\
\midrule
OLMo-2 1B / G & 0.83 & 0.67 & 0.83 & 0.83 \\
OLMo-2 1B / C & 0.83 & 0.83 & 0.83 & 0.67 \\
OLMo-2 1B / M & 0.83 & 0.33 & 0.67 & 0.33 \\
SmolLM2 / G & 0.90 & 0.80 & 0.80 & 0.30 \\
SmolLM2 / C & 0.80 & 0.60 & 0.70 & 0.60 \\
SmolLM2 / M & 0.80 & 0.50 & 0.90 & 0.60 \\
\bottomrule
\end{tabular}

\end{table}

\section{Cross-domain transfer and successive changes}
\label{app:fields}
Table~\ref{tab:domaintransfer} lists the nine leave-one-domain-out correlations. Each template averages the depth-standardized endpoint change vectors of the other two domains. In the trimmed comparison, both measured endpoints are removed before standardization. The correlations assess change shape and leave the held-out domain's absolute scale unspecified. For OLMo-2 1B and SmolLM2, each paired-window bootstrap draw reconstructs the source vectors and the held-out target before computing the correlation.

We additionally remove regional mean changes before computing pairwise correlations. Here the measured positions are divided into three contiguous groups whose sizes differ by at most one, assigning any extra positions to earlier groups. The resulting median correlation is 0.947. This comparison indicates that a shared coarse regional offset is insufficient to account for the cross-domain agreement. Its grouping serves a different purpose from the fixed-third geography in Appendix~\ref{app:profiles}, and can differ at a boundary by one block.

\begin{table}[!htb]
\caption{\textbf{Leave-one-domain-out redistribution patterns.} All comparisons hold the trajectory and training endpoints fixed. Trimming removes the first and last measured blocks before constructing the template.}
\label{tab:domaintransfer}
\centering\small\begin{tabular}{@{}llrr@{}}
\toprule
Trajectory & Held-out text & All measured & Trimmed ends \\
\midrule
OLMo-2 1B & Code & 0.875 & 0.881 \\
OLMo-2 1B & General & 0.933 & 0.935 \\
OLMo-2 1B & Math & 0.944 & 0.918 \\
OLMo-2 7B & Code & 0.934 & 0.920 \\
OLMo-2 7B & General & 0.944 & 0.962 \\
OLMo-2 7B & Math & 0.986 & 0.959 \\
SmolLM2 & Code & 0.993 & 0.990 \\
SmolLM2 & General & 0.979 & 0.986 \\
SmolLM2 & Math & 0.993 & 0.995 \\
\bottomrule
\end{tabular}

\end{table}

\begin{figure}[!htb]
\centering
\includegraphics[width=\linewidth]{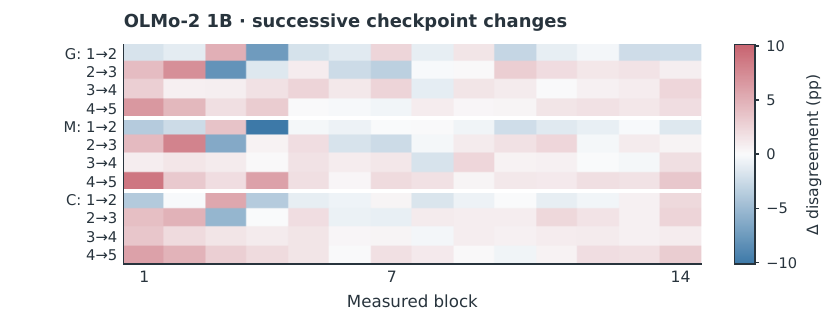}\par
\includegraphics[width=\linewidth]{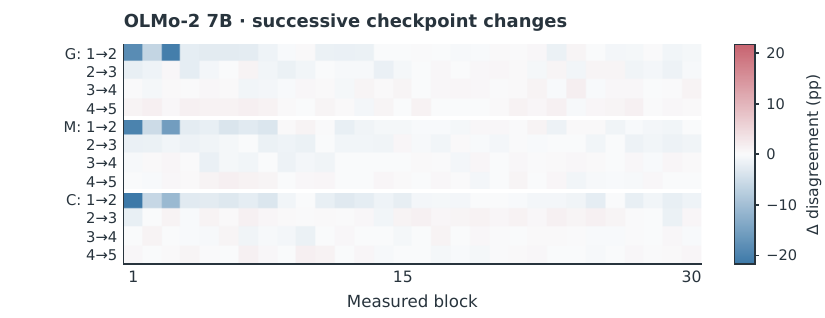}\par
\includegraphics[width=\linewidth]{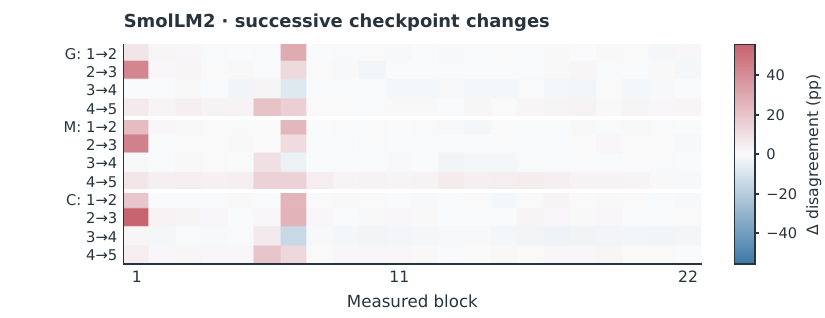}
\caption{\textbf{Successive sensitivity changes across text domains.} G/M/C denote general, mathematical, and code text. Rows show adjacent-checkpoint differences, and columns show measured interior blocks. Red indicates an increase and blue a decrease. Each model has its own symmetric color scale. The recurring depth patterns coexist with differences in change magnitude.}
\label{fig:fields}
\end{figure}

Figure~\ref{fig:fields} displays the four successive checkpoint changes for each domain. Vectorizing the interval-by-depth matrices gives median within-trajectory correlation 0.900 across nine domain pairs, or 0.861 after removing both measured endpoints. This extends the endpoint observation to shared locations of change within the sampled training intervals. The intervals cover unequal numbers of training tokens, so the heatmaps represent differences over each interval rather than rates per token.

\section{Continuous-output metrics and numerical validation}
\label{app:continuous}
\paragraph{Output metrics.}
For intact logits $z$ and bypassed logits $z'$, let $p=\operatorname{softmax}(z)$ and $q=\operatorname{softmax}(z')$. We use natural logarithms for forward KL, $\operatorname{KL}(p\|q)$, and JS (Eq.~\ref{eq:js}). Target-token NLL change is $-\log q_y+\log p_y$. For the intact top token $a$ and runner-up $b$, margin loss is $(z_a-z_b)-(z'_a-z'_b)$, holding the intact competitor fixed. Positive NLL change denotes a lower probability for the observed target; positive margin loss denotes reduced separation of the intact top token from its original competitor.

Table~\ref{tab:endpoints} reports all five endpoint changes. Pythia increases across metrics and OLMo decreases, showing that their opposite trends occur in both discrete decisions and continuous output measures. Table~\ref{tab:continuousranks} compares depth ordering. KL, JS, and target NLL retain stronger rank correspondence than margin loss, illustrating that the chosen output observable also affects longitudinal persistence.

\begin{table}[!htb]
\caption{\textbf{Continuous and discrete response changes.} Pythia intervals resample six paired training runs. OLMo contributes one trajectory. Disagreement is in probability units; multiplying by 100 gives percentage points.}
\label{tab:endpoints}
\centering\small\begin{tabular}{@{}lrrr@{}}
\toprule
Endpoint & Pythia change & 95\% run interval & OLMo change \\
\midrule
Top-1 disagreement & +0.0792 & [0.0677, 0.0881] & -0.0146 \\
Forward KL & +0.2492 & [0.2194, 0.2722] & -0.0269 \\
Jensen--Shannon & +0.0493 & [0.0435, 0.0537] & -0.0053 \\
Target NLL change & +0.2498 & [0.2234, 0.2710] & -0.0313 \\
Logit-margin loss & +0.1066 & [0.0792, 0.1292] & -0.0274 \\
\bottomrule
\end{tabular}

\end{table}

\begin{table}[!htb]
\caption{\textbf{Depth-rank persistence across output metrics.} Values are early--final correlations of the continuous-output experiment's layer profiles.}
\label{tab:continuousranks}
\centering\small\begin{tabular}{@{}lrr@{}}
\toprule
Endpoint & Pythia $\rho$ & OLMo $\rho$ \\
\midrule
Top-1 disagreement & 0.903 & 0.785 \\
Forward KL & 0.891 & 0.787 \\
Jensen--Shannon & 0.915 & 0.766 \\
Target NLL change & 0.879 & 0.705 \\
Logit-margin loss & 0.430 & 0.547 \\
\bottomrule
\end{tabular}

\end{table}

\paragraph{Fixed-margin comparisons.}
Pythia's pooled early intact top-versus-runner-up margin cutpoints are 0.296875, 0.6953125, 1.28125, and 2.4140625. The same cutpoints define five strata at both checkpoints. Disagreement, KL, JS, and NLL changes have positive intervals in all five strata; margin loss is positive in four. A preregistered regression adjustment retains a positive checkpoint coefficient for disagreement (0.4706 log-odds), conditional on intact margin and the specified model terms. The stratified comparisons are the main check that the training effect persists at comparable margins; regression coefficients and leave-one-run-out results are available in the supplement.

\paragraph{Numerical reproducibility.}
The initial cross-environment Pythia evaluation did not reproduce historical integer disagreement counts exactly. Before fresh outcome acquisition, a documented protocol revision replaced exact cross-environment count equality with a representation-derived ambiguity bound, deterministic repeats, and reference-input checks before fresh measurements were collected. All 30 historical layer/checkpoint counts lie within the resulting admissible intervals. OLMo separately reproduces its historical counts exactly. Both evaluations pass checks for finite logits, normalized probabilities, deterministic inference, identity interventions, and fixed token counts. Continuous metrics are computed in FP32 from FP16 model states.

The fraction of Pythia positions eligible for ambiguity under the representation bound increases from 0.01813 to 0.31089. This is a bound-based eligibility measure, rather than an observed frequency of incorrect predictions. Historical logits were not archived, so the identities of exact historical argmax changes cannot be recovered. The supplementary protocol includes the amendment and numerical checks needed to interpret the cross-environment comparison.

\section{Matched local perturbations and downstream response}
\label{app:matching}
\paragraph{Protocol sequence.}
The continuous-output audit precedes representation-aware magnitude matching and direction controls. Matched results use fresh measurements under the representation-aware protocol; earlier halted pilot measurements are excluded. The supplement provides the pre-analysis protocols, revisions and numerical validation records.

\paragraph{Perturbation construction and precision.}
For each paired token and block, Eq.~\ref{eq:match} defines the common relative target $r^*$. With $\beta^*=r^*/r_{\rm nat}$, we construct the ideal perturbed residual in FP32 and round it to the inference precision:
\begin{equation}
 h_{\rm ideal}=\operatorname{FP32}(h^+)+\beta^*[\operatorname{FP32}(h)-\operatorname{FP32}(h^+)],
 \qquad h_{\rm exec}=\operatorname{FP16}(h_{\rm ideal}).
\end{equation}
The measured perturbation is $\operatorname{FP32}(h_{\rm exec})-\operatorname{FP32}(h^+)$. Its relative RMS, $r_{\rm exec}$, uses the intact-state denominator in Eq.~\ref{eq:match}. The zero-perturbation and full-bypass endpoints return $h^+$ and $h$ exactly. Matched comparisons exclude paired observations with natural relative magnitude below $10^{-4}$ or intact-state RMS below $10^{-6}$, applying the exclusions symmetrically across checkpoints.

The representation error bound is $\RMS(\operatorname{FP32}(h_{\rm exec})-h_{\rm ideal})/(\RMS(h^+)+10^{-8})$. Ideal target error must be at most $\max(10^{-7},10^{-5}|r^*|)$. Observed target error, less ideal error and the representation bound, must be at most $5\times10^{-7}$. All 14 model/checkpoint evaluations pass these checks. Measuring the post-rounding difference ensures that the analysis controls the perturbation actually presented to the suffix.

\paragraph{Adjustment for actual relative magnitude.}
Within each Pythia run, the preregistered least-squares model predicts matched JS using checkpoint, standardized $z=\log(r_{\rm exec}+10^{-12})$, $z^2$, normalized depth and its square, target-scale indicators, and checkpoint interactions with $z$ and $z^2$. Predictions under early and final checkpoint assignments are averaged over the same retained covariate distribution. The six adjusted run-level contrasts are resampled 100,000 times with seed 20260922. This marginal standardization compares output response at a common distribution of measured perturbation magnitudes and depths.

OLMo uses the same adjustment with a 10,000-window bootstrap, seed 20260923. Its adjusted matched-JS interval is $[-0.00023059,-0.00021978]$, whereas the unadjusted window interval includes zero. These intervals describe within-trajectory variation under their respective estimators. Table~\ref{tab:scales} reports Pythia's scale-specific contrasts. The positive small-scale and negative larger-scale changes explain why the pooled estimate must be interpreted over the four specified target scales.

\begin{table}[!htb]
\caption{\textbf{Pythia matched-JS changes by target scale.} The primary adjusted comparison pools the four scales; these scale-specific comparisons reveal their different checkpoint trends.}
\label{tab:scales}
\centering\small\begin{tabular}{@{}rrr@{}}
\toprule
$\gamma$ & Matched-JS change & 95\% run interval \\
\midrule
0.25 & +0.001291 & [+0.000866, +0.001784] \\
0.50 & +0.000114 & [-0.000739, +0.000945] \\
0.75 & -0.003289 & [-0.005309, -0.001642] \\
1.00 & -0.008606 & [-0.014081, -0.004576] \\
\bottomrule
\end{tabular}

\end{table}

\begin{figure}[!htb]
\centering\includegraphics[width=.58\linewidth]{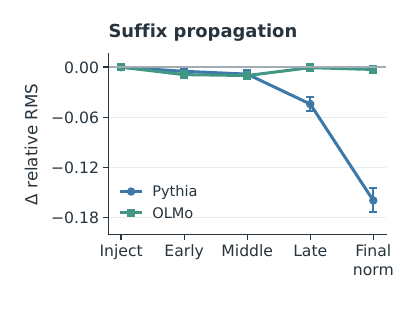}
\caption{\textbf{Downstream perturbation changes accumulate late in the suffix.} Final-minus-early contrasts in hidden-state relative RMS at injection, successive suffix regions, and final normalization. Pythia intervals resample paired training runs; OLMo contributes one trajectory. The increasingly negative Pythia contrast indicates a larger checkpoint difference near the output boundary.}
\label{fig:propagation}
\end{figure}

\paragraph{Propagation over depth.}
The injection point defines the initial perturbation. We divide the remaining Transformer boundaries by relative suffix progress into $(0,1/3]$, $(1/3,2/3]$, and $(2/3,1]$, followed by final normalization. Figure~\ref{fig:propagation} reports final-minus-early contrasts in relative perturbation magnitude at these boundaries. The Pythia contrast grows in magnitude toward the end of the suffix and reaches $-0.1597$ at final normalization. The direction-control experiment separately measures the incremental change caused by normalization.

\paragraph{Held-out prediction of continuous response.}
We hold out windows 2 and 5 from each text-sampling seed, giving six test windows and 12 training windows. The split is shared across runs, checkpoints, blocks, and intervention conditions. The target is $\log(\JS+10^{-12})$. Every regression includes normalized depth, squared depth, and Pythia run indicators. The magnitude model adds $\log(r_{\rm exec}+10^{-12})$; the propagation model adds the log ratio of final-normalization to injected relative perturbation magnitude. The joint model includes both, and the checkpoint-aware model adds checkpoint and its interactions with both features. Standardization is fitted on training windows only.

Table~\ref{tab:pred} shows that magnitude predicts substantial held-out variation, while propagation and residual checkpoint terms provide additional information. The joint score exceeds the preregistered adequacy threshold of $R^2=0.20$. These predictive comparisons complement the matched contrasts by showing that a single measured component leaves systematic output variation unexplained.

\begin{table}[!htb]
\caption{\textbf{Held-out prediction of log JS.} Models use local magnitude, downstream propagation, or both. The final model adds checkpoint-dependent terms. OLMo scores describe one trajectory.}
\label{tab:pred}
\centering\small\begin{tabular}{@{}lrr@{}}
\toprule
Predictors & Pythia test $R^2$ & OLMo test $R^2$ \\
\midrule
Local magnitude & 0.514 & 0.394 \\
Downstream propagation & 0.015 & 0.074 \\
Magnitude + propagation & 0.543 & 0.457 \\
Joint + checkpoint terms & 0.623 & 0.456 \\
\bottomrule
\end{tabular}

\end{table}

\section{Direction-dependent propagation}
\label{app:directions}
\paragraph{Controls.}
We evaluate Pythia blocks 1, 5, and 10 and OLMo blocks 1, 15, and 30. Four Gaussian directions, using seeds 3101--3104, are projected orthogonally to the natural missing-update direction at each token. The base random draws are shared across paired checkpoints. A second control assigns to token position $p$ the natural direction at $(p+137)\bmod256$ within the same window and block. Controls are scaled to the natural perturbation's ideal absolute RMS and measured after FP16 rounding. Exclusions and representation-error checks follow Appendix~\ref{app:matching}; absolute cosines with the natural direction must be at most $10^{-5}$ before rounding and 0.01 after rounding for the orthogonal controls.

\begin{figure}[!htb]
\centering\twoplots{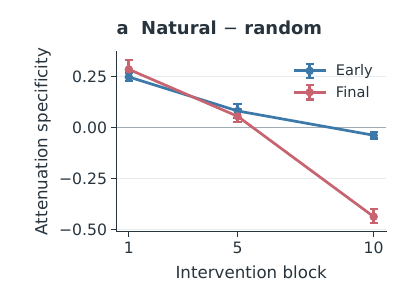}{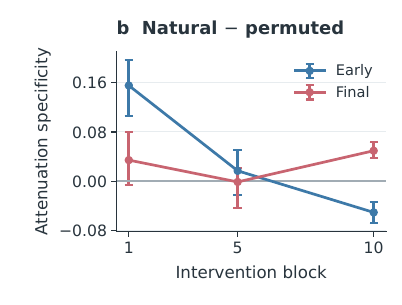}
\caption{\textbf{Directional specificity varies with depth and control family.} Early and final Pythia contrasts use paired-run intervals. Negative values indicate stronger attenuation of the natural direction. Random-orthogonal and token-permuted controls give different depth patterns. Neither control comparison establishes targeted compensation.}
\label{fig:directions}
\end{figure}

\paragraph{Attenuation and specificity.}
For direction $d$, define $A_d=\log(r_{d,\rm finalnorm}/r_{d,\rm initial})$. Specificity is the natural-minus-control contrast in $A_d$, adjusted for actual norm and projection onto the intact top-versus-runner-up unembedding contrast. Negative values indicate stronger natural-direction attenuation. Figure~\ref{fig:directions} and Table~\ref{tab:specificity} show that the final aggregate is negative relative to random directions and positive relative to permuted directions. Preferential attenuation therefore depends on the control family.

\begin{table}[!htb]
\caption{\textbf{Adjusted Pythia directional specificity.} Checkpoint changes are paired within the same runs and control constructions.}
\label{tab:specificity}
\centering\small\begin{tabular}{@{}lrrrr@{}}
\toprule
Contrast & Early & Final & Change & 95\% change interval \\
\midrule
Natural $-$ random & +0.0966 & -0.0328 & -0.1294 & [-0.1450, -0.1125] \\
Natural $-$ permuted & +0.0406 & +0.0276 & -0.0130 & [-0.0318, +0.0066] \\
\bottomrule
\end{tabular}

\end{table}

\paragraph{Preregistered criterion for targeted compensation.}
At either checkpoint, both control-family contrasts must have a negative mean of magnitude at least 0.05 with an interval excluding zero. Each must be negative in at least five of six runs and have negative upper interval bounds in at least two of three layers. Both must satisfy the protocol's downstream-localization requirements. Direction identity must add at least 0.01 held-out $R^2$, with a positive interval, after readout and generic-propagation adjustment. Neither checkpoint meets this joint criterion; the results do not identify targeted natural-direction compensation.

\paragraph{Predictive comparisons.}
On the held-out windows from Appendix~\ref{app:matching}, additional predictors are evaluated after magnitude, depth, checkpoint, and run terms. Generic propagation adds 0.10146 in $R^2$; readout projection and direction identity add 0.00232 and 0.00783 (Table~\ref{tab:directionprediction}). Propagation is informative, with direction-dependent structure remaining unresolved.

\begin{table}[!htb]
\caption{\textbf{Held-out information from directional controls.} Intervals resample the six Pythia runs. The preregistered threshold for a material increment is 0.01 in $R^2$.}
\label{tab:directionprediction}
\centering\small\begin{tabular}{@{}lrr@{}}
\toprule
Added information & Test $R^2$ increment & 95\% run interval \\
\midrule
Readout projection & 0.00232 & [0.00189, 0.00277] \\
Generic propagation & 0.10146 & [0.09061, 0.11245] \\
Direction-family identity & 0.00783 & [0.00606, 0.00915] \\
\bottomrule
\end{tabular}

\end{table}

At final block 10, normalization changes specificity by $-0.5602$ relative to random controls and $-0.0669$ relative to permuted controls, yet total post-normalization specificity against permuted controls remains positive (+0.0495). A large normalization effect thus coexists with weak overall natural-direction selectivity. None of the three explanatory accounts meets its full preregistered criterion.

\clearpage
\section{Historical replication and released response fields}\label{app:dense}
\begin{figure}[p]
\centering\includegraphics[width=\linewidth]{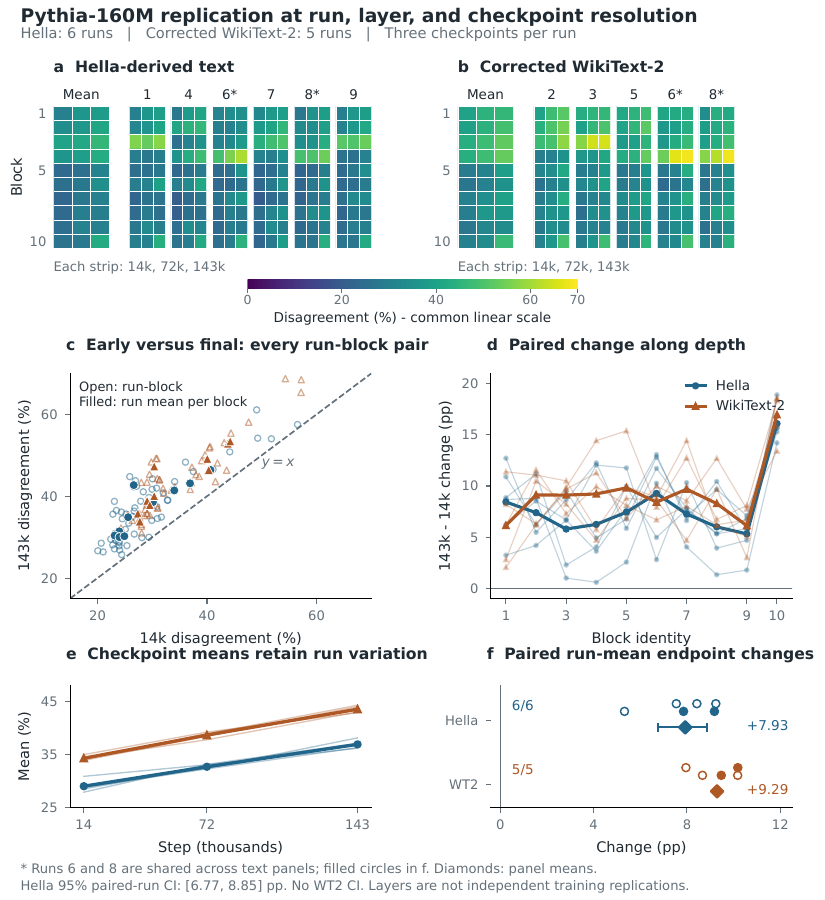}
\caption{\textbf{Historical Pythia replication at run, layer and checkpoint resolution.} (a,b) All 330 run--block--checkpoint cells use a common 0--70\% scale. Asterisks identify shared runs 6 and 8. (c) Open points show 110 run--block endpoint pairs; filled points are run-averaged layers. (d,e) Thin traces retain runs and bold traces show panel means. (f) All six Hella and five corrected WikiText-2 run-mean changes are positive, averaging $+7.93$ and $+9.29$ pp. The Hella whisker is the retained 95\% paired-run interval $[6.77,8.85]$ pp; no WikiText interval is introduced. Layers and repeated displays are not independent training replications.}
\label{fig:dense-replication}
\end{figure}

The historical Hella and corrected WikiText-2 panels in Figure~\ref{fig:dense-replication} are separate from the amended continuous-output execution in Figure~\ref{fig:continuous}. Their six and five runs share identities 6 and 8. The historical Hella lowest-response pair changes from $\{5,7\}$ to $\{8,9\}$ despite endpoint rank correlation $0.867$. Reusing the early pair at the final checkpoint costs $0.79$ percentage points, with the retained paired-run interval $[0,2.23]$. This surrogate averages separate single-block bypasses, not joint pruning or task loss. The saved bootstrap reselects both pairs inside each run resample; no new resampling was performed. The original tie policy and complete environment lockfile are not available.

\end{document}